\documentclass[sigchi]{acmart}

\usepackage{bm}
\usepackage{multirow}

\AtBeginDocument{%
  \providecommand\BibTeX{{%
    \normalfont B\kern-0.5em{\scshape i\kern-0.25em b}\kern-0.8em\TeX}}}

\setcopyright{rightsretained}
\copyrightyear{2020}
\acmYear{2020}

\acmConference[IUI '20 Workshops]{}{March 17, 2020}{Cagliari, Italy}
\acmBooktitle{IUI '20 Wokrshops, March 17, 2020, Cagliari, Italy}

\begin{document}

\title{A Multi-Turn Emotionally Engaging Dialog Model}

\author{Yubo Xie}
\email{yubo.xie@epfl.ch}
\author{Ekaterina Svikhnushina}
\email{ekaterina.svikhnushina@epfl.ch}
\author{Pearl Pu}
\email{pearl.pu@epfl.ch}
\affiliation{%
  \institution{\'{E}cole Polytechnique F\'{e}d\'{e}rale de Lausanne}
  \city{Lausanne}
  \country{Switzerland}
}

\renewcommand{\shortauthors}{Xie, et al.}

\begin{abstract}
  Open-domain dialog systems (also known as chatbots) have increasingly drawn attention in natural language processing. Some of the recent work aims at incorporating affect information into sequence-to-sequence neural dialog modeling, making the response emotionally richer, while others use hand-crafted rules to determine the desired emotion response. However, they do not explicitly learn the subtle emotional interactions captured in human dialogs. In this paper, we propose a multi-turn dialog system aimed at learning and generating emotional responses that so far only humans know how to do. Compared with two baseline models, offline experiments show that our method performs the best in perplexity scores. Further human evaluations confirm that our chatbot can keep track of the conversation context and generate emotionally more appropriate responses while performing equally well on grammar.
\end{abstract}



\maketitle

\section{Introduction}

Many application areas show significant benefits of integrating affect information in natural language dialogs. In earlier work on human computer interaction, Klein et al.~\cite{DBLP:journals/iwc/KleinMP01} found user's frustration caused by a computer system can be alleviated by computer-initiated emotional support, by providing feedback on emotional content along with sympathy and empathy. Recently, Hu et al.~\cite{DBLP:conf/chi/HuXLYGSLA18} developed a customer support neural chatbot, capable of generating dialogs similar to the humans in terms of empathic and passionate tones, potentially serving as proxy customer support agents on social media platforms. In a qualitative study~\cite{DBLP:conf/hai/Zamora17}, participants expressed an interest in chatbots capable of serving as an attentive listener and providing motivational support, thus fulfilling users' emotional needs. Several participants even noted a chatbot is ideal for sensitive content that is too embarrassing to ask another human. Finally Bickmore and Picard~\cite{DBLP:journals/tochi/BickmoreP05} showed a relational agent with deliberate social-emotional skills was respected more, liked more, and trusted more, even after four weeks of interaction, compared to an equivalent task-oriented agent.

Recent development in neural language modeling has generated significant excitement in the open-domain dialog generation community. The success of sequence-to-sequence (seq2seq) learning~\cite{DBLP:conf/nips/SutskeverVL14,DBLP:conf/emnlp/ChoMGBBSB14} in the field of neural machine translation has inspired researchers to apply the recurrent neural network (RNN) encoder-decoder structure to response generation~\cite{DBLP:journals/corr/VinyalsL15}. Following the standard seq2seq structure, various improvements have been made on the neural conversation model. For example, Shang et al.~\cite{DBLP:conf/acl/ShangLL15} applied attention mechanism~\cite{DBLP:journals/corr/BahdanauCB14} to the same structure on Twitter-style microblogging data. Li et al.~\cite{DBLP:conf/naacl/LiGBGD16} found the original version tend to favor short and dull responses. They fixed this problem by increasing the diversity of the response. Li et al.~\cite{DBLP:conf/acl/LiGBSGD16} modeled the personalities of the speakers, and Xing et al.~\cite{DBLP:conf/aaai/XingWWLHZM17} developed a topic aware dialog system. We call work in this area globally neural dialog generation. For a comprehensive survey, please refer to~\cite{DBLP:journals/sigkdd/ChenLYT17}.

More recently, researchers started incorporating affect information
into neural dialog models. While a central theme seems to be making the responses emotionally richer, existing approaches mainly follow two directions. In one, an emotion label is explicitly required as input so that the machine can generate sentences of that particular emotion label or type~\cite{DBLP:conf/aaai/ZhouHZZL18}. In another group of work, the main idea is to develop handcrafted rules to direct the machines to generated responses of the desired emotions~\cite{DBLP:conf/ecir/AsgharPHJM18,DBLP:conf/aaai/Zhong0M19}.  Both approaches require an emotion label as input (either given or handcrafted), which might be unpractical in real dialog scenarios.

Furthermore, to the best of our knowledge, the psychology and social science literature does not provide clear rules for emotional interaction. It seems such social and emotional intelligence is captured in our conversations. This is why we decided to take the automatic and data-driven approach. In this paper, we describe an end-to-end Multi-turn Emotionally Engaging Dialog model (MEED), capable of recognizing emotions and generating emotionally appropriate and human-like responses with the ultimate goal of reproducing social behaviors that are habitual in human-human conversations. We chose the multi-turn setting because a model suitable for single-turn dialogs cannot effectively track earlier context in multi-turn dialogs, both semantically and emotionally. Since being able to track several turns is really important, we made this design decision from the beginning, in contrast to most related work where models are only trained and tested on single-turn dialogs. While using a hierarchical mechanism to track the conversation history in multi-turn dialogs is not new (e.g., HRAN by Xing et al.~\cite{DBLP:conf/aaai/XingWWHZ18}), to combine it with an additional emotion RNN to process the emotional information in each history utterance has never been attempted before.

Our contributions are threefold. (1) We describe in detail a novel emotion-tracking dialog generation model that learns the emotional interactions directly from the data. This approach is free of human-defined heuristic rules, and hence, is more robust and fundamental than those described in existing work. (2) We compare our model, MEED, with the generic seq2seq model and the hierarchical model of multi-turn dialogs (HRAN). Offline experiments show that our model outperforms both seq2seq and HRAN by a significant amount. Further experiments with human evaluation show our model produces emotionally more appropriate responses than both baselines, while also improving the language fluency. (3) We illustrate a human-evaluation procedure for judging machine produced emotional dialogs. We consider factors such as the balance of positive and negative emotions in test dialogs, a well-chosen range of topics, and dialogs that our human evaluators can relate. It is the first time such an approach is designed with consideration for human judges. Our main goal is to increase the objectivity of the results and reduce judges' mistakes due to out-of-context dialogs they have to evaluate.


\section{Related Work}
\label{sec:related_work}

\subsection{Neural Dialog Generation}
Vinyals and Le~\cite{DBLP:journals/corr/VinyalsL15} were one of the first to model dialog generation using neural networks. Their seq2seq framework was trained on an IT Helpdesk Troubleshooting dataset and the OpenSubtitles dataset~\cite{DBLP:conf/lrec/LisonT16}. Shang et al.~\cite{DBLP:conf/acl/ShangLL15} further trained the seq2seq model with attention mechanism on a self-crawled Weibo (a popular Twitter-like social media website in China) dataset. Meanwhile, Xu et al.~\cite{DBLP:conf/chi/XuLGSA17} built a customer service chatbot by training the seq2seq model on a dataset collected with conversations between customers and customer service accounts from 62 brands on Twitter.

The standard seq2seq framework is applied to single-turn response generation. In multi-turn settings, where a context with multiple history utterances is given, the same structure often ignores the hierarchical characteristic of the context. Some recent work addresses this problem by adopting a hierarchical recurrent encoder-decoder (HRED) structure~\cite{DBLP:conf/cikm/SordoniBVLSN15,DBLP:conf/aaai/SerbanSBCP16,DBLP:conf/aaai/SerbanSLCPCB17}. To give attention to different parts of the context while generating responses, Xing et al.~\cite{DBLP:conf/aaai/XingWWHZ18} proposed the hierarchical recurrent attention network (HRAN), using a hierarchical attention mechanism. However, these multi-turn dialog models do not take into account the turn-taking emotional changes of the dialog.

\subsection{Neural Dialog Models with Affect Information}

Recent work on incorporating affect information into natural language processing tasks has inspired our current work. They can be mainly described as affect language models and emotional dialog systems.

Ghosh et al.~\cite{DBLP:conf/acl/GhoshCLMS17} made the first attempt to augment the original LSTM language model with affect treatment in what they called Affect-LM. At training time, Affect-LM can be considered as an energy based model where the added energy term captures the degree of correlation between the next word and the affect information of the preceeding text. At text generation time, affect information is also used to increase the appropriate selection of the next word. A key component in Affect-LM is the use of a well established text analysis program, LIWC (Linguistic Inquiry and Word Count)~\cite{pennebaker2001linguistic}. For every sentence, for example, ``I unfortunately did not pass my exam'', the model generates five emotion features denoting (\textit{sad}:\,1, \textit{angry}:\,1, \textit{anxiety}:\,1, \textit{negative emotion}:\,1, \textit{positive emotion}:\,0). This makes Affect-LM both capable of distinguishing affect information conveyed by each word in the language modeling part and aware of the preceeding text's emotion in each generation step. In a similar vein, Asghar et al.~\cite{DBLP:conf/ecir/AsgharPHJM18} appended the original word embeddings with a VAD affect model~\cite{warriner2013norms}. VAD is a vector model, as opposed to a categorical model (LIWC), representing a given emotion in each of the valence, arousal, and dominance axes. In contrast to Affect-LM, Asghar's neural affect dialog model aims at generating explicit responses given a particular utterance. To do so, the authors designed three affect-related loss functions, namely minimizing affect dissonance, maximizing a affective dissonance, and maximizing affective content. The paper also proposed the affectively diverse beam search during decoding, so that the generated candidate responses are as affectively diverse as possible. However, literature in affective science does not necessarily validate such rules. In fact, the best strategy to
speak to an angry customer is the de-escalation strategy (using
neutral words to validate anger) rather than employing equally emotional words (minimizing affect dissonance) or words that convey happiness (maximizing affect dissonance).

The Emotional Chatting Machine (ECM)~\cite{DBLP:conf/aaai/ZhouHZZL18} takes a post and generates a response in a predefined emotion category. The main idea is to use an internal memory module to capture the emotion dynamics during decoding, and an external memory module to model emotional expressions explicitly by assigning different probability values to emotional words as opposed to regular words. Zhou and Wang~\cite{DBLP:conf/acl/WangZ18} extended the standard seq2seq model to a conditional variational autoencoder combined with policy gradient techniques. The model takes a post and an emoji as input, and generates the response with target emotion specified by the emoji. Hu et al.~\cite{DBLP:conf/chi/HuXLYGSLA18} built a tone-aware chatbot for customer care on social media, by deploying extra meta information of the conversations in the seq2seq model. Specifically, a tone indicator is added to each step of the decoder during the training phase. 

In parallel to these developments, Zhong et al.~\cite{DBLP:conf/aaai/Zhong0M19} proposed an affect-rich dialog model using biased attention mechanism on emotional words in the input message, by taking advantage of the VAD embeddings. The model is trained with a weighted cross-entropy loss function, which encourages the generation of emotional words.

\subsection{Summary}

As much as these work in the above section inspired our work, our approach in generating affect dialogs is significantly different. Most of related work focused on integrating affect information into the transduction vector space using either VAD or LIWC, we aim at modeling and generating the affect exchanges in human dialogs using a dedicated embedding layer. The approach is also completely data-driven, thus absent of hand-crafted rules. To avoid learning obscene and callous exchanges often found in social media data like tweets and Reddit threads~\cite{DBLP:conf/acl/RashkinSLB19}, we opted to train our model on movie subtitles, whose dialogs were carefully created by professional writers. We believe the quality of this dataset can be better than those curated by crowdsource platforms. For modeling the affect information, we chose to use LIWC because it is a well-established emotion lexical resource, covering the whole English dictionary whereas VAD only contains 13K lemmatized terms.

\section{Model}
\label{sec:model}

\begin{figure*}[t]
    \centering
    \includegraphics{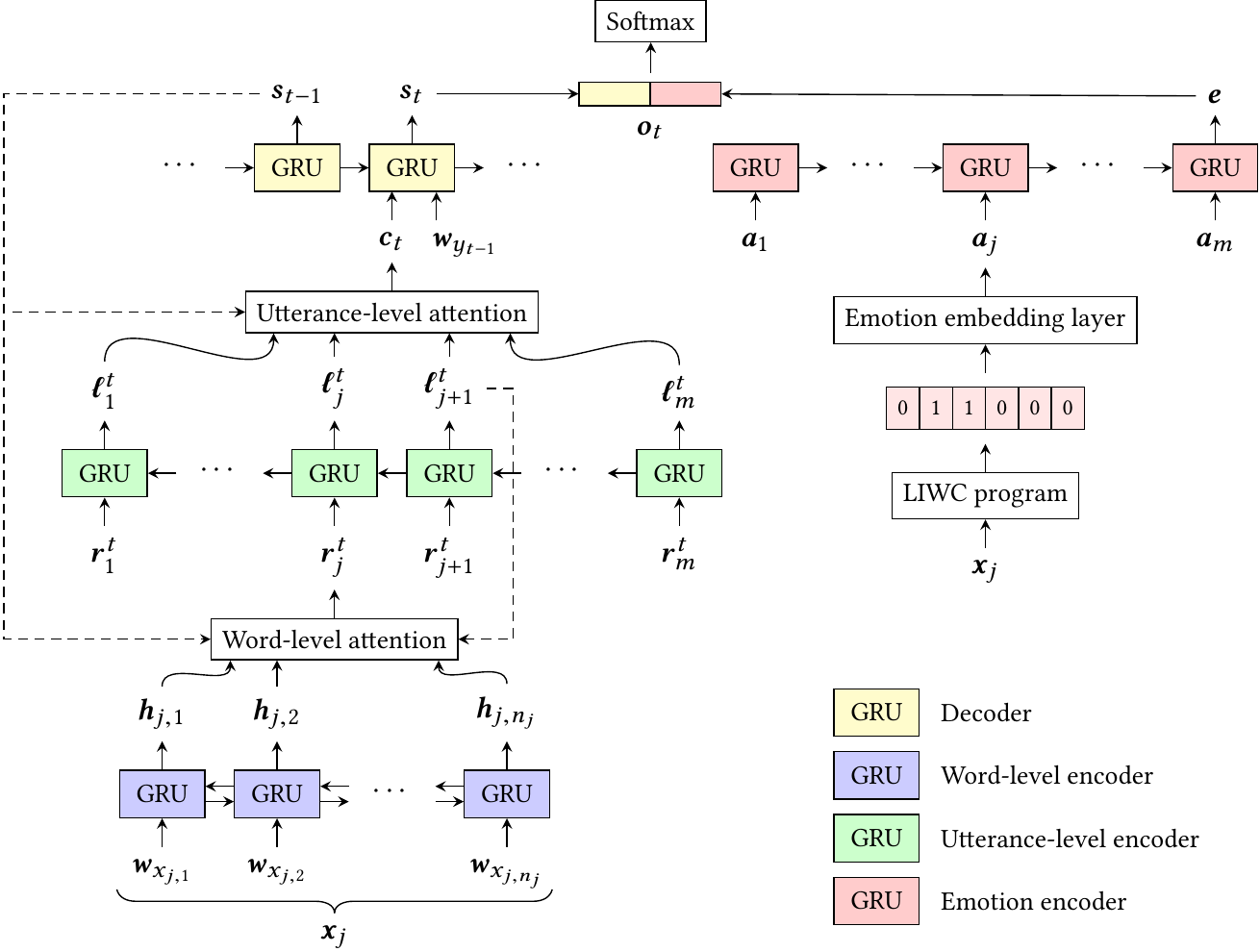}
    \caption{The overall architecture of our model.}
    \label{fig:model}
    \Description{The overall architecture of our model. The model consists of two major parts. On the left part, we have the hierarchical attention mechanism that encodes the input utterances into a context vector. On the right part, we have an emotion encoder that recognizes the emotion states from input and encodes them into one emotion vector. These two vectors are then concatenated and sent into a softmax layer to produce tokens in the response.}
\end{figure*}

We describe our model one element at a time, from the basic structure, to the hierarchical component, and finally the emotion embedding layer.

We first consider the problem of generating response $\bm{y}$ given a context $\bm{X}$ consisting of multiple previous utterances by estimating the probability distribution $p(\bm{y}\,|\,\bm{X})$ from a data set $\mathcal{D}=\{(\bm{X}^{(i)},\bm{y}^{(i)})\}_{i=1}^N$ containing $N$ context-response pairs. Here
\begin{equation}
    \bm{X}^{(i)} = \big(\bm{x}_1^{(i)},\bm{x}_2^{(i)},\dots,\bm{x}_{m_i}^{(i)}\big)
\end{equation}
is a sequence of $m_i$ utterances, and
\begin{equation}
    \bm{x}_j^{(i)} = \big(x^{(i)}_{j,1},x^{(i)}_{j,2},\dots,x^{(i)}_{j,n_{ij}}\big)
\end{equation}
is a sequence of $n_{ij}$ words. Similarly,
\begin{equation}
    \bm{y}^{(i)} = \big(y_{1}^{(i)},y_{2}^{(i)},\dots,y_{T_i}^{(i)}\big)
\end{equation}
is the response with $T_i$ words.

Usually the probability distribution $p(\bm{y}\,|\,\bm{X})$ can be modeled by an RNN language model conditioned on $\bm{X}$. When generating the word $y_t$ at time step $t$, the context $\bm{X}$ is encoded into a fixed-sized dialog context vector $\bm{c}_t$ by following the hierarchical attention structure in HRAN~\cite{DBLP:conf/aaai/XingWWHZ18}. Additionally, we extract the emotion information from the utterances in $\bm{X}$ by leveraging an external text analysis program, and use an RNN to encode it into an emotion context vector $\bm{e}$, which is combined with $\bm{c}_t$ to produce the distribution. The overall architecture of the model is depicted in Figure~\ref{fig:model}. We are going to elaborate on how to obtain $\bm{c}_t$ and $\bm{e}$, and how they are combined in the decoding part.

\subsection{Hierarchical Attention}

The hierarchical attention structure involves two encoders to produce the dialog context vector $\bm{c}_t$, namely the word-level encoder and the utterance-level encoder. The word-level encoder is essentially a bidirectional RNN with gated recurrent units (GRU)~\cite{DBLP:conf/emnlp/ChoMGBBSB14}. For utterance $\bm{x}_j$ in $\bm{X}$ ($j=1,2,\dots,m$), the bidirectional encoder produces two hidden states at each word position $k$, the forward hidden state $\bm{h}^\mathrm{f}_{jk}$ and the backward hidden state $\bm{h}^\mathrm{b}_{jk}$. The final hidden state $\bm{h}_{jk}$ is then obtained by concatenating the two,
\begin{equation}
    \bm{h}_{jk} = \mathrm{concat}\big(\bm{h}^\mathrm{f}_{jk}, \bm{h}^\mathrm{b}_{jk}\big).
\end{equation}
The utterance-level encoder is a unidirectional RNN with GRU that goes from the last utterance in the context to the first, with its input at each step as the summary of the corresponding utterance, which is obtained by applying a Bahdanau-style attention mechanism~\cite{DBLP:journals/corr/BahdanauCB14} on the word-level encoder output. More specifically, at decoding step $t$, the summary of utterance $\bm{x}_j$ is a linear combination of $\bm{h}_{jk}$, for $k=1,2,\dots,n_j$,
\begin{equation}
    \bm{r}_j^t = \sum_{k=1}^{n_j} \alpha_{jk}^t \bm{h}_{jk}.
\end{equation}
Here $\alpha_{jk}^t$ is the word-level attention score placed on $\bm{h}_{jk}$, and can be calculated as
\begin{align}
a_{jk}^t &= \bm{v}_a^T \tanh(\bm{U}_a\bm{s}_{t-1} + \bm{V}_a\bm{\ell}_{j+1}^t + \bm{W}_a\bm{h}_{jk}), \\
\alpha_{jk}^t &= \frac{\exp(a_{jk}^t)}{\sum_{k^\prime=1}^{n_j}\exp(a_{jk^\prime}^t)},
\end{align}
where $\bm{s}_{t-1}$ is the previous hidden state of the decoder, $\bm{\ell}_{j+1}^t$ is the previous hidden state of the utterance-level encoder, and $\bm{v}_a$, $\bm{U}_a$, $\bm{V}_a$ and $\bm{W}_a$ are word-level attention parameters. The final dialog context vector $\bm{c}_t$ is then obtained as another linear combination of the outputs of the utterance-level encoder $\bm{\ell}_{j}^t$, for $j=1,2,\dots,m$,
\begin{equation}
    \bm{c}_t = \sum_{j=1}^{m} \beta_{j}^t \bm{\ell}_{j}^t.
\end{equation}
Here $\beta_{j}^t$ is the utterance-level attention score placed on $\bm{\ell}_{j}^t$, and can be calculated as
\begin{align}
b_{j}^t &= \bm{v}_b^T \tanh(\bm{U}_b\bm{s}_{t-1} + \bm{W}_b\bm{\ell}_{j}^t), \\
\beta_{j}^t &= \frac{\exp(b_{j}^t)}{\sum_{j^\prime=1}^{m}\exp(b_{j^\prime}^t)},
\end{align}
where $\bm{s}_{t-1}$ is the previous hidden state of the decoder, and $\bm{v}_b$, $\bm{U}_b$ and $\bm{W}_b$ are utterance-level attention parameters.

\subsection{Emotion Encoder}
The main objective of the emotion embedding layer is to recognize the affect information in the given utterances so that the model can respond with emotionally appropriate replies. To achieve this, we need an encoder to distinguish the affect information in the context, in addition to its semantic meaning. Equally we need a decoder capable of selecting the best and most human-like answers.

We are able to achieve this goal, i.e., capturing the emotion information carried in the context $\bm{X}$, in the encoder, thanks to LIWC. We make use of the five emotion-related categories, namely \emph{positive emotion}, \emph{negative emotion}, \emph{anxious}, \emph{angry}, and \emph{sad}. This set can be expanded to include more categories if we desire a richer distinction. See the discussion section for more details on how to do this. Using the newest version of the program LIWC2015,\footnote{\url{https://liwc.wpengine.com/}} we are able to map each utterance $\bm{x}_j$ in the context to a six-dimensional indicator vector $\bm{1}(\bm{x}_j)$, with the first five entries corresponding to the five emotion categories, and the last one corresponding to \emph{neutral}. If any word in $\bm{x}_j$ belongs to one of the five categories, then the corresponding entry in $\bm{1}(\bm{x}_j)$ is set to $1$; otherwise, $\bm{x}_j$ is treated as neutral, with the last entry of $\bm{1}(\bm{x}_j)$ set to $1$. For example, assuming $\bm{x}_j=$ ``he is worried about me'', then
\begin{equation}
    \bm{1}(\bm{x}_j) = [0, 1, 1, 0, 0, 0],
\end{equation}
since the word ``worried'' is assigned to both \emph{negative emotion} and \emph{anxious}. We apply a dense layer with sigmoid activation function on top of $\bm{1}(\bm{x}_j)$ to embed the emotion indicator vector into a continuous space,
\begin{equation}
    \bm{a}_j = \sigma(\bm{W}_e\bm{1}(\bm{x}_j) + \bm{b}_e),
\end{equation}
where $\bm{W}_e$ and $\bm{b}_e$ are trainable parameters. The emotion flow of the context $\bm{X}$ is then modeled by an unidirectional RNN with GRU going from the first utterance in the context to the last, with its input being $\bm{a}_j$ at each step. The final emotion context vector $\bm{e}$ is obtained as the last hidden state of this emotion encoding RNN.

\subsection{Decoding}

The probability distribution $p(\bm{y}\,|\,\bm{X})$ can be written as
\begin{align}
p(\bm{y}\,|\,\bm{X}) &= p(y_1,y_2,\dots,y_T\,|\,\bm{X}) \nonumber \\
&= p(y_1\,|\,\bm{c}_1,\bm{e}) \prod_{t=2}^T p(y_t\,|\,y_1,\dots,y_{t-1},\bm{c}_t,\bm{e}).
\label{eq:prob_dist}
\end{align}
We model the probability distribution using an RNN language model along with the emotion context vector $\bm{e}$. Specifically, at time step $t$, the hidden state of the decoder $\bm{s}_t$ is obtained by applying the GRU function,
\begin{equation}
    \bm{s}_t = \mathrm{GRU}(\bm{s}_{t-1},\mathrm{concat}(\bm{c}_t,\bm{w}_{y_{t-1}})),
\end{equation}
where $\bm{w}_{y_{t-1}}$ is the word embedding of $y_{t-1}$. Similar to Affect-LM~\cite{DBLP:conf/acl/GhoshCLMS17}, we then define a new feature vector $\bm{o}_t$ by concatenating $\bm{s}_t$ (which we refer to as the language context vector) with the emotion context vector $\bm{e}$,
\begin{equation}
    \bm{o}_t = \mathrm{concat}(\bm{s}_t,\bm{e}),
\end{equation}
on which we apply a softmax layer to obtain a probability distribution over the vocabulary,
\begin{equation}
    \bm{p}_t = \mathrm{softmax}(\bm{W}\bm{o}_t + \bm{b}),
    \label{eq:softmax}
\end{equation}
where $\bm{W}$ and $\bm{b}$ are trainable parameters.
Each term in Equation~(\ref{eq:prob_dist}) is then given by
\begin{equation}
    p(y_t\,|\,y_1,\dots,y_{t-1},\bm{c}_t,\bm{e}) = \bm{p}_{t,y_t}.
\end{equation}
We use the cross-entropy loss as our objective function
\begin{equation}
    \mathcal{L} = -\frac{1}{\sum_{i=1}^N T_i} \sum_{i=1}^N \log p\big( \bm{y}^{(i)}\,|\,\bm{X}^{(i)} \big).
\end{equation}

\section{Evaluation}
\label{sec:evaluation}

We trained our model using two different datasets and compared its performance with HRAN as well as the basic seq2seq model by performing both offline and online testings.

\subsection{Datasets}
We used two different dialog corpora to train our model---the Cornell Movie Dialogs Corpus~\cite{DBLP:conf/acl-cmcl/Danescu-Niculescu-Mizil11} and the DailyDialog dataset~\cite{li2017dailydialog}.
\begin{itemize}
\item \textbf{Cornell Movie Dialogs Corpus}. The dataset contains 83,097 dialogs (220,579 conversational exchanges) extracted from raw movie scripts. In total there are 304,713 utterances.
\item \textbf{DailyDialog}. The dataset is developed by crawling raw data from websites used for language learners to learn English dialogs in daily life. It contains 13,118 dialogs in total.
\end{itemize}
We summarize some of the basic information regarding the two datasets in Table~\ref{tab:datasets}.

\begin{table}[t]
\centering
\caption{Statistics of the two datasets.}
\begin{tabular}{lrr}
\toprule
& Cornell & DailyDialog \\
\midrule
\# dialogs & 83,097 & 13,118 \\
\# utterances & 304,713 & 102,977 \\
Average \# turns & 3.7 & 7.9 \\
Average \# words / utterance & 12.5 & 14.6 \\
\midrule
Training set size & 142,450 & 46,797 \\
Validation set size & 10,240 & 10,240 \\
\bottomrule
\end{tabular}
\label{tab:datasets}
\end{table}

In our experiments, the models were first trained on the Cornell Movie Dialogs Corpus, and then fine-tuned on the DailyDialog dataset. We adopted this training pattern because the Cornell dataset is bigger but noisier, while DailyDialog is smaller but more daily-based. To create a training set and a validation set for each of the two datasets, we took segments of each dialog with number of turns no more than six,\footnote{We chose the maximum number of turns to be six because we would like to have a longer context for each dialog while at the same time keeping the training procedure computationally efficient.} to serve as the training/validation examples. Specifically, for each dialog $\bm{D}=(\bm{x}_1,\bm{x}_2,\dots,\bm{x}_M)$, we created $M-1$ context-response pairs, namely $\bm{U}_i=(\bm{x}_{s_i},\dots,\bm{x}_i)$ and $\bm{y}_i=\bm{x}_{i+1}$, for $i=1,2,\dots,M-1$, where $s_i=\max(1,i-4)$. We filtered out those pairs that have at least one utterance with length greater than 30. We also reduced the frequency of those pairs whose responses appear too many times (the threshold is set to 10 for Cornell, and 5 for DailyDialog), to prevent them from dominating the learning procedure. See Table~\ref{tab:datasets} for the sizes of the training and validation sets. The test set consists of 100 dialogs with four turns. We give more detailed description of how we created the test set in the section of human evaluation.

\subsection{Baselines and Implementation}
Our choice of including S2S is rather obvious. Including HRAN instead of other neural dialog models with affect information was not an easy decision. As mentioned in the related work, Asghar's affective dialog model, the affect-rich conversation model, and the Emotional Chatting Machine do not learn the emotional exchanges in the dialogs. This leaves us wondering whether using a multi-turn neural model can be as effective in learning emotional exchanges as MEED. In addition, comparing S2S and HRAN also gives us an idea of how much the hierarchical mechansim is improving upon the basic model. This is why our final comparision is based on three multi-turn dialog generation models: the standard seq2seq model (denoted as S2S), HRAN, and our proposed model, MEED. In order to adapt S2S to the multi-turn setting, we concatenate all the history utterances in the context into one.

For all the models, the vocabulary consists of 20,000 most frequent words in the Cornell and DailyDialog datasets, plus three extra tokens: \texttt{<unk>} for words that do not exist in the vocabulary, \texttt{<go>} indicating the begin of an utterance, and \texttt{<eos>} indicating the end of an utterance. Here we summarize the configurations and parameters of our experiments:
\begin{itemize}
    \item We set the word embedding size to 256. We initialized the word embeddings in the models with word2vec~\cite{DBLP:journals/corr/abs-1301-3781} vectors first trained on Cornell and then fine-tuned on DailyDialog, consistent with the training procedure of the models.
    \item We set the number of hidden units of each RNN to 256, the word-level attention depth to 256, and utterance-level 128. The output size of the emotion embedding layer is 256.
    \item We optimized the objective function using the Adam optimizer~\cite{DBLP:journals/corr/KingmaB14} with an initial learning rate of 0.001.
    \item For prediction, we used beam search~\cite{DBLP:journals/coling/TillmannN03} with a beam width of 256.
\end{itemize}
We have made the source code publicly available.\footnote{\url{https://github.com/yuboxie/meed}}

\subsection{Evaluation Metrics}
\label{subsec:metrics}

The evaluation of chatbots remains an open problem in the field. Recent work~\cite{DBLP:conf/emnlp/LiuLSNCP16} has shown that the automatic evaluation metrics borrowed from machine translation such as BLEU score~\cite{DBLP:conf/acl/PapineniRWZ02} tend to align poorly with human judgement. Therefore, in this paper, we mainly adopt human evaluation, along with perplexity and BLEU score, following the existing work.

\subsubsection{Automatic Evaluation}
Perplexity is a measurement of how a probability model predicts a sample. It is a popular method used in language modeling. In neural dialog generation community, many researchers have adopted this method, especially in the beginning of this field~\cite{DBLP:journals/corr/VinyalsL15,DBLP:conf/aaai/SerbanSBCP16,DBLP:conf/aaai/XingWWHZ18,DBLP:conf/acl/WangZ18,DBLP:conf/aaai/ZhouHZZL18,DBLP:conf/aaai/Zhong0M19}. It measures how well a dialog model predicts the target response. Given a target response $\bm{y}=\{y_1,y_2,\dots,y_T\}$, the perplexity is calculated as
\begin{align}
    \text{ppl}(\bm{y}) &= p(y_1,y_2,\dots,y_T)^{-1/T} \nonumber \\
    &= \exp{\Bigg[ -\frac{1}{T}\sum_{t=1}^T \log{p(y_t\,|\,y_1,\dots,y_{t-1})} \Bigg]}.
\end{align}
Thus a lower perplexity score indicates that the model has better capability of predicting the target sentence, i.e., the humans' response. Some researchers~\cite{DBLP:conf/acl/ShangLL15,DBLP:conf/emnlp/LiMRJGG16,DBLP:conf/aaai/Zhong0M19} argue that perplexity score is not the ideal measurement because for a given context history, one should allow many responses. This is especially true if we want our conversational agents to speak more diversely. However, for our purpose, which is to speak emotionally appropriately and as human-like as possible, we believe this is a good measure. We do recognize that it is not the only way to measure chatbots' performance. This is why we also conducted human evaluation experiment.

BLEU score is often used to measure the quality of machine-translated text. Some earlier work of dialog response generation~\cite{DBLP:conf/naacl/LiGBGD16,DBLP:conf/acl/LiGBSGD16} adopted this metric to measure the performance of chatbots. However, recent study~\cite{DBLP:conf/emnlp/LiuLSNCP16} suggests that it does not align well with human evaluation. Nevertheless, we still include BLEU scores in this paper, to get a sense of comparison with perplexity and human evaluation results.

\subsubsection{Human Evaluation}

Human evaluation has been widely used to evaluate open-domain dialog generation tasks. This approach can include any criterion as we judge appropriate. Most commonly, researchers have included the model's ability to generate grammatically correct, contextually coherent, and emotionally appropriate responses, of which the latter two properties cannot be reliably evaluated using automatic metrics. Recent work~\cite{DBLP:conf/ecir/AsgharPHJM18, DBLP:conf/aaai/Zhong0M19, DBLP:conf/aaai/ZhouHZZL18} on affect-rich conversational chatbots turned to human opinion to evaluate both fluency and emotionality of their models. But such human experiments are sensitive to risk factors if the experiment is not carefully designed. They include whether the intructions are clear, whether they have been tested with users before hand, and whether there is a good balance of the human judgement tasks. Further, if a test set for human evaluation is prepared by randomly sampling the dialogs from the dataset, it may include out-of-context dialogs, causing confusion and ambiguity for human evaluators. Unbalanced emotional distribution of the test dialogs may also lead to biased conclusions since the chatbot's abilities are evaluated on the unrepresentative sample.

To take into account the above issues, we took several iterations to prepare the instructions and the test set before conducting the human evaluation experiment. Part of our test set comes from the DailyDialog dataset, which consists of meaningful complete dialogs. To compensate for the inbalance, we further curated more negative emotion dialogs so that the final set has equal emotion distributions. We provide the details about the test data preparation process and the evaluation experiment below.

\paragraph{Preparation of Natural Dialog Test Set}

We first selected the emotionally colored dialogs with exactly four turns from the DailyDialog dataset.
In the dataset each dialog turn is annotated with a corresponding emotional category, including the neutral one. For our purposes we filtered out only those dialogs where more than a half of utterances have non-neutral emotional labels, resulting in 78 emotionally positive dialogs and 14 emotionally negative dialogs. We recruited two human workers to augment the data to produce more emotionally negative dialogs. Both of them were PhD students from our university (males, aged 24 and 25), fluent in English, and not related to the authors' lab. We found them via email and messaging platforms, and offered 80 CHF (or roughly US \$80) gift coupons as incentive for each participant. The workers fulfilled the tasks in Google form\footnote{We provide the link to the form used for creating the dialogs: \url{https://forms.gle/rPagMZYuYJ3M3Sq8A}, hoping to help other researchers reproduce the same procedure. However, due to privacy concerns, we do not plan to release this dataset.} following the instructions and created five negative dialogs with four turns, as if they were interacting with another human, in each of the following topics: \emph{relationships}, \emph{entertainment}, \emph{service}, \emph{work and study}, and \emph{everyday situations}. The Google form was released on 31 January 2019, and the workers finished their tasks by 4 February 2019. Subsequently, to form the final test set, we randomly selected 50 emotionally positive and 50 emotionally negative dialogs from the two pools of dialogs described above.

\paragraph{Human Evaluation Experiment Design}

In the final human evaluation of the model, we recruited four more PhD students from our university (1 female and 3 males, aged 22--25). Three of them are fluent English speakers and one is a native speaker. The recruitment proceeded in the same manner as described above; the raters were offered 80 CHF (or roughly US \$80) per participant gift coupons for fulfilling the task, and extra 20 CHF (or roughly US \$20) coupon was promised as a bonus to the rater judged to be the most serious. For the evaluation survey, we also leveraged Google form. Specifically, we randomly shuffled the 100 dialogs in the test set, then we used the first three utterances of each dialog as the input to the three models being compared (S2S, HRAN, and MEED), and obtain the respective responses. Dialog contexts and three models' responses were included into Google form. According to the context given, the raters were instructed to evaluate the quality of the responses based on three criteria:
\begin{enumerate}
    \item \emph{Grammatical correctness}---whether or not the response is fluent and free of grammatical mistakes;
    \item \emph{Contextual coherence}---whether or not the response is context sensitive to the previous dialog history;
    \item \emph{Emotional appropriateness}---whether or not the response conveys the right emotion and feels as if it had been produced by a human.
\end{enumerate}
For each criterion, the raters gave scores of either 0, 1 or 2, where 0 means bad, 2 means good, and 1 indicates neutral.
For this survey, the Google form was launched on 12 February 2019, and all the submissions from our raters were collected by 14 February 2019.

\subsection{Results and Analysis}

In this subsection, we present the experimental results of the automatic evaluation metric as well as human judgement, followed by some analysis.

\subsubsection{Automatic Evaluation Results}

\begin{table*}[t]
\centering
\caption{Perplexity and average BLEU scores achieved by the models. Avg.~BLEU: average of BLEU-1, -2, -3, and -4. Validation set~1 comes from the Cornell dataset, and validation set 2 comes from the DailyDialog dataset.}
\begin{tabular}{lcccccc}
\toprule
& \multicolumn{3}{c}{Perplexity} & \multicolumn{3}{c}{Avg.~BLEU} \\
\cmidrule(lr){2-4}
\cmidrule(lr){5-7}
& Validation Set 1 & Validation Set 2 & Test Set & Validation Set 1 & Validation Set 2 & Test Set \\
\midrule
S2S & 43.136 & 25.418 & 19.913 & 1.639 & 2.427 & 3.720 \\
HRAN & 46.225 & 26.338 & 20.355 & 1.701 & 2.368 & 2.390 \\
MEED & \textbf{41.862} & \textbf{24.341} & \textbf{19.795} & \textbf{1.829} & \textbf{2.635} & \textbf{4.281} \\
\bottomrule
\end{tabular}
\label{tab:perplexity}
\end{table*}

Table~\ref{tab:perplexity} gives the perplexity and BLEU scores obtained by the three models on the two validation sets and the test set. As shown in the table, MEED achieves the lowest perplexity and the highest BLEU score on all three sets. We conducted $t$-test on the perplexity obtained, and results show significant improvements of MEED over S2S and HRAN on the two validation sets (with $p$-value $<0.05$).

\subsubsection{Human Evaluation Results}

\begin{table}[t]
\centering
\caption{Human evaluation results on grammatical correctness.}
\begin{tabular}{lccccc}
\toprule
& +2 & +1 & 0 & Avg.~Score & $r$ \\
\midrule
S2S & 98.0 & 0.8 & 1.2 & 1.968 & 0.915 \\
HRAN & 98.5 & 1.3 & 0.2 & 1.982 & 0.967 \\
MEED  & 99.5 & 0.3  & 0.2  & \textbf{1.992} & 0.981 \\
\bottomrule
\end{tabular}
\label{tab:human_grammar}
\end{table}
\begin{table}[t]
\centering
\caption{Human evaluation results on contextual coherence.}
\begin{tabular}{lccccc}
\toprule
& +2 & +1 & 0 & Avg.~Score & $\kappa$ \\
\midrule
S2S & 25.8 & 19.7 & 54.5 & 0.713 & 0.389 \\
HRAN & 37.3 & 21.2 & 41.5 & 0.958 & 0.327 \\
MEED  & 38.5 & 22.0  & 39.5  & \textbf{0.990} & 0.356 \\
\bottomrule
\end{tabular}
\label{tab:human_context}
\end{table}
\begin{table}[t]
\centering
\caption{Human evaluation results on emotional appropriateness.}
\begin{tabular}{lccccc}
\toprule
& +2 & +1 & 0 & Avg.~Score & $\kappa$ \\
\midrule
S2S & 21.8 & 25.2 & 53.0 & 0.688 & 0.361 \\
HRAN & 30.5 & 28.5 & 41.0 & 0.895 & 0.387 \\
MEED  & 32.0 & 27.8  & 40.2  & \textbf{0.917} & 0.337 \\
\bottomrule
\end{tabular}
\label{tab:human_emotion}
\end{table}

Table~\ref{tab:human_grammar}, \ref{tab:human_context} and \ref{tab:human_emotion} summarize the human evaluation results on the responses' grammatical correctness, contextual coherence, and emotional appropriateness, respectively. In the tables, we give the percentage of votes each model received for the three scores, the average score obtained, and the agreement score among the raters. Note that we report Fleiss' $\kappa$ score~\cite{fleiss1973equivalence} for contextual coherence and emotional appropriateness, and Finn's $r$ score~\cite{finn1970note} for grammatical correctness.
We did not use Fleiss' $\kappa$ score for grammatical correctness. As agreement is extremely high, this can make Fleiss' $\kappa$ very sensitive to prevalence~\cite{DBLP:journals/jbi/HripcsakH02}. 
On the contrary, we did not use Finn's $r$ score for contextual coherence and emotional appropriateness because it is only reasonable when the observed variance is significantly less than the chance variance~\cite{tinsley1975interrater}, which did not apply to these two criteria. As shown in the tables, we got high agreement among the raters for grammatical correctness, and fair agreement among the raters for contextual coherence and emotional appropriateness.\footnote{\url{https://en.wikipedia.org/wiki/Fleiss\%27_kappa\#Interpretation}} For grammatical correctness, all three models achieved high scores, which means all models are capable of generating fluent utterances that make sense. For contextual coherence and emotional appropriateness, MEED achieved higher average scores than S2S and HRAN, which means MEED keeps better track of the context and can generate responses that are emotionally more appropriate and natural. We first conducted Friedman test~\cite{howell2016fundamental} and then $t$-test on the human evaluation results (contextual coherence and emotional appropriateness), showing the improvements of MEED over S2S are significant (with $p$-value $<0.01$).

The comparison between perplexity scores and human evaluation results further confirms the fact that in the context of dialog response generation, perplexity does not align with human judgement. In Table~\ref{tab:perplexity}, for all the three sets, HRAN performs worse than S2S in terms of perplexity. However, for all of the three criteria in human evaluation, HRAN actually outperforms S2S. Based on this, we conclude that perplexity alone is not enough for evaluating a dialog system.

\subsubsection{Visualization of Output Layer Weights}

\begin{figure*}[t]
    \centering
    \includegraphics[width=0.3\textwidth]{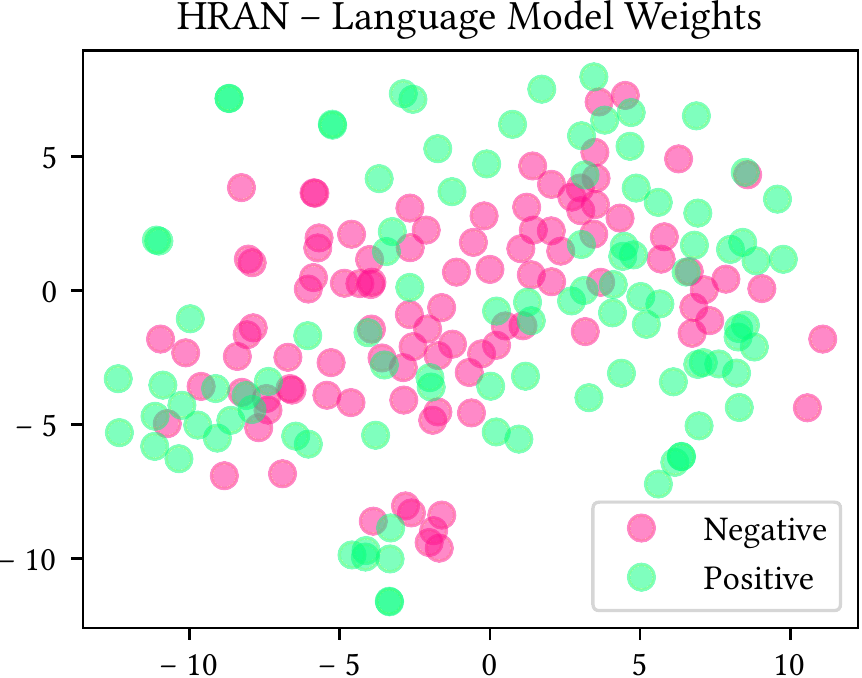}\qquad
    \includegraphics[width=0.3\textwidth]{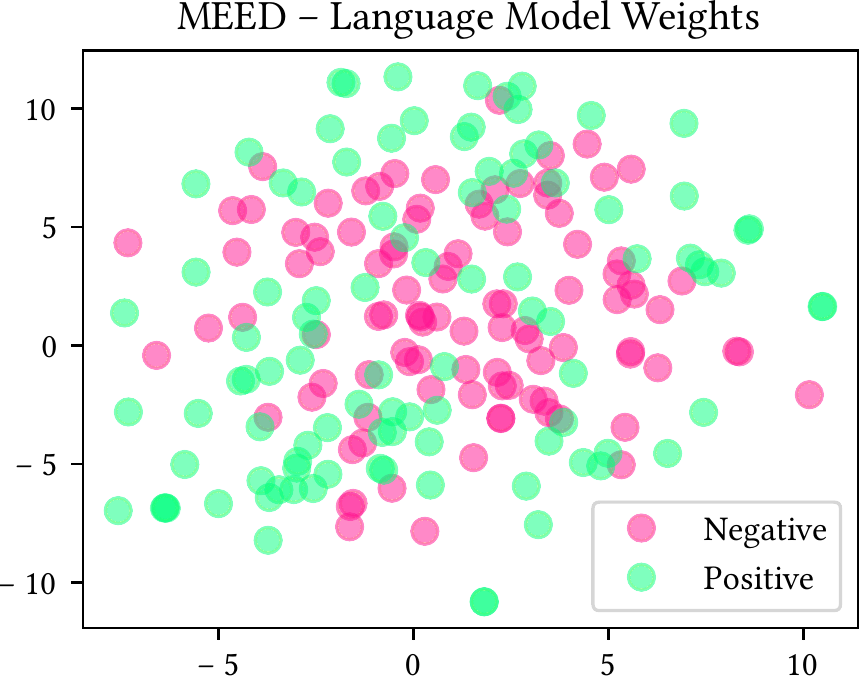}\qquad
    \includegraphics[width=0.3\textwidth]{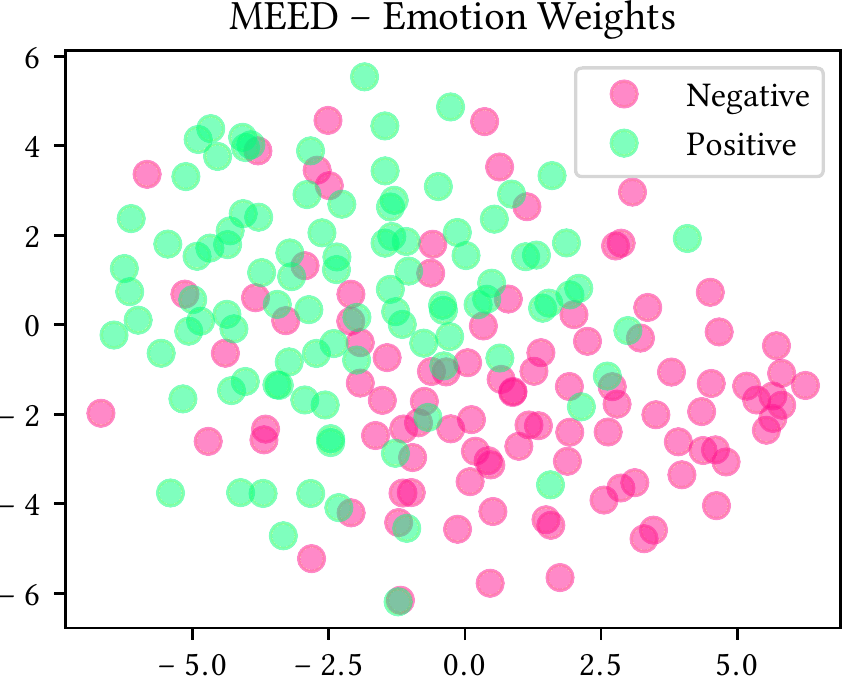}
    \caption{t-SNE visualization of the output layer weights in HRAN and MEED. 100 most frequent positive words and 100 most frequent negative words are shown. The weight vectors in MEED are separated into two parts and visualized individually.}
    \label{fig:visual}
    \Description{t-SNE visualization of the output layer weights in HRAN and MEED. 100 most frequent positive words and 100 most frequent negative words are shown. The weight vectors in MEED are separated into two parts and visualized individually. The first two plots visualize the language model weights of HRAN and MEED respectively, where positive words (green dots) and negative words (red dots) are mixed with each other. The last plot visualizes the emotion weights of MEED, where positive words are mainly grouped on the top-left, and negative words are mainly grouped at the bottom-right.}
\end{figure*}

We may wonder how HRAN and MEED differ in terms of the distributional representations of their respective vocabularies (words in the language model, and affect words). We decided to visualize the output layer weights as word embedding representations using dimensionality reduction technique for the various models. 

In the decoding phase, Equation~(\ref{eq:softmax}) takes $\bm{o}_t$, the concatenation of the language context vector $\bm{s}_t$ and the emotion context vector $\bm{e}$, and generates a probability distribution over the vocabulary words by applying a softmax layer. The weight matrix of this softmax layer is denoted as $\bm{W}$, whose shape is $|V|\times 2d$, where $|V|$ is the vocabulary size and $d=256$ is the hidden state size of the RNNs. Thus the $i$th row of the weight matrix $\bm{W}_i$ can be regarded as a vector representation of the $i$th word in the vocabulary. Since we concatenate the language context vector and the emotion context vector as the input to the softmax layer, the first half of the weight vector $\bm{W}_i$ corresponds to the language context vector, and the second half corresponds to the emotion context vector. We refer to them as language model weights and emotion weights, respectively. If the emotion embedding layer is learning and distinguishing affect states correctly, we will see clear differences in the visualization.

With t-SNE~\cite{maaten2008visualizing}, we are able to reduce the dimensionality of the weights to two, and visualize them in a straightforward way. For better illustration, we selected 100 most frequent (emotionally) positive words and 100 most frequent negative words from the vocabulary, and used t-SNE to project the corresponding language model weights and emotion weights to two dimensions. Figure~\ref{fig:visual} gives the results in three subplots. Since HRAN does not have the emotion context vector, we just visualized the whole output layer weight vector, which does a similar job as the language model weights in MEED. We can observe from the first two plots that positive words (green dots) and negative words (red dots) are scattered around and mixed with each other in the language model weights for HRAN and MEED respectively, which means no emotion information is captured in these weights. On the contrary, the emotion weights in MEED, in the last plot, have a clearer clustering effect, i.e., positive words are mainly grouped on the top-left, while negative words are mainly grouped at the bottom-right. This gives the hint that the emotion encoder in MEED is capable of tracking the emotion states in the conversation history.

\subsubsection{Case Study}

\begin{table*}[t]
\centering
\caption{Sample model responses. For each dialog, the ground truth is included in a pair of parentheses.}
\begin{tabular}{cll}
\toprule
& Context & Model Responses \\
\midrule
\multirow{4}{*}{1} & A: I'm happy to see you again. & \multirow{4}{*}{\parbox{4.15cm}{S2S: I hope so.\\HRAN: Thanks a lot.\\MEED: That sounds like fun.}} \\
& B: Mee too. & \\
& A: We should do this more often. & \\
& (B: Okay, I'll give you a ring next week.) & \\
\midrule
\multirow{4}{*}{2} & A: Thank god! I am finished writing that service guide! It took me forever! & \multirow{4}{*}{\parbox{4.15cm}{S2S: When?\\HRAN: Why?\\MEED: Congratulations!}} \\
& B: When did you finish? & \\
& A: This morning! No more overtime, and no more headaches! & \\
& (B: Well, I'm glad to hear it. Have a cup of coffee!) & \\
\midrule
\multirow{4}{*}{3} & A: I think that's settled. & \multirow{4}{*}{\parbox{4.15cm}{S2S: What is it?\\HRAN: What is it?\\MEED: Are you serious?}} \\
& B: I'm tired of your cut-and-dried opinions. Who do you think you are! & \\
& A: How dare you speak to me like this. & \\
& (B: Why not?) & \\
\midrule
\multirow{4}{*}{4} & A: This concert was awful. & \multirow{4}{*}{\parbox{4.15cm}{S2S: Congratulations!\\HRAN: Why not?\\MEED: That's true.}} \\
& B: Agreed, the musicians were not in harmony. & \\
& A: It was too painful. Never again. & \\
& (B: That's for sure!) & \\
\bottomrule
\end{tabular}
\label{tab:examples}
\end{table*}

We present four sample dialogs in Table~\ref{tab:examples}, along with the responses generated by the three models. Dialog~1 and 2 are emotionally positive and dialog~3 and 4 are negative. For the first two examples, we can see that MEED is able to generate more emotional content (like ``fun'' and ``congratulations'') that is appropriate according to the context. For dialog 4, MEED responds in sympathy to the other speaker, which is consistent with the second utterance in the context. On the contrary, HRAN poses a question in reply, contradicting the dialog history.

\section{Discussion}
\label{sec:discussion}

In this section, we briefly discuss how our framework can incorporate other components, as well as several directions to extend it.

\subsection{Emotion Recognition}

To extract the affect information contained in the utterances, we used the LIWC text analysis program. We believe this emotion recognition step is vital for a dialog model to produce emotionally appropriate responses. However, the choice of emotion classifier is not strictly limited to LIWC. It could be replaced by other well-established affect recognizer or one that is more appropriate to the target domain. For example, we can consider using more fine-grained emotion categories from GALC~\cite{scherer2005emotions}, or using DeepMoji~\cite{DBLP:conf/emnlp/FelboMSRL17}, which was trained on millions of tweets with emoji labels and is more suitable for tweet-like conversations. However, for DeepMoji, the 64 categories of emojis do not have a clear and exact correspondence with standardized emotion categories, nor to the VAD vectors.

\subsection{Training Data}

We pre-trained our model on the Cornell movie subtitles and then fine-tuned it with the DailyDialog dataset. We adopted this particular training order because we would like our chatbot to talk more like human chit-chats, and the DailyDialog dataset, compared with the bigger Cornell dataset, is more daily-based. Since our model learns how to respond properly in a data-driven way, we believe having a training dataset with good quality while being large enough plays an important role in developing an engaging and user-friendly chatbot. Thus, in the future, we plan to train our model on the multi-turn conversations that we have already extracted from the much bigger OpenSubtitles corpus and the EmpatheticDialogues dataset.\footnote{\url{https://github.com/facebookresearch/EmpatheticDialogues}}

\subsection{Evaluation}

Evaluation of dialog models remains an open problem in the response generation field. Early work~\cite{DBLP:conf/emnlp/RitterCD11,DBLP:conf/naacl/SordoniGABJMNGD15,DBLP:conf/acl/LiGBSGD16} on response generation used automatic evaluation metrics borrowed from the machine translation field, such as the BLEU score, to evaluate dialog systems. Later on, Liu et al.~\cite{DBLP:conf/emnlp/LiuLSNCP16} showed that these metrics correlate poorly with human judgement. Recently, a number of researchers begain developing automatic and data-driven evaluation methods~\cite{DBLP:conf/acl/LoweNSABP17,DBLP:conf/aaai/TaoMZY18}, with the ultimate goal of replacing human evaluation. However they are still in an early stage. In this paper, we used both perplexity measures and human judgement in our experiments to finalize our model. In other words, using the perplexity measures, we were able to determine when to stop training our model. But this condition does not gurantee the optimal results until human judgement test can validate them. We thus highly recommend this combination, which is also a common practice in the research community~\cite{DBLP:conf/aaai/XingWWHZ18,DBLP:conf/acl/WangZ18,DBLP:conf/aaai/ZhouHZZL18,DBLP:conf/aaai/Zhong0M19}. 

\subsection{Model Extensions}

Our model uses RNNs to encode the input sequences, and GRU cells to capture long-term dependency among different positions in the sequences. Recent advances in natural language understanding have proposed new network architectures to process text input. Specifically, the Transformer~\cite{DBLP:conf/nips/VaswaniSPUJGKP17} uses pure attention mechanisms without any recurrence structures. Compared with RNNs, the Transformer can capture better long-term dependency due to the self-attention mechanism, which is free of locality biases, and is more efficient to train because of better parallelization capability. Following the Transformer architecture, researchers found that pre-training language models on huge amounts of data could largely boost the performance of downstream tasks, and published many pre-trained language models such as BERT~\cite{DBLP:conf/naacl/DevlinCLT19} and RoBERTa~\cite{DBLP:journals/corr/abs-1907-11692}. As future work, we would like to adopt the Transformer architecture to replace the RNNs in our model, and initialize our encoder with pre-trained language models. We hope to increase the performance of response generation.

\section{Conclusion}
\label{sec:conclusion}

 We believe reproducing conversational and emotional intelligence will make social chatbots more believable and engaging. In this paper, we proposed a multi-turn dialog system capable of recognizing and generating emotionally appropriate responses, which is the first step toward such a goal. We have demonstrated how to do so by (1) modeling utterances with extra affect vectors, (2) creating an emotional encoding mechanism that learns emotion exchanges in the dataset, (3) curating a multi-turn and balanced dialog dataset, and (4) evaluating the model with offline and online experiments. For future directions, we would like to investigate the diversity issue of the responses generated, possibly by extending the mutual information objective function~\cite{DBLP:conf/naacl/LiGBGD16} to multi-turn settings. We would also like to adopt the Transformer architecture with pre-trained language model weights, and train our model on a much larger dataset, by extracting multi-turn dialogs from the OpenSubtitles corpus.


\bibliographystyle{ACM-Reference-Format}
\bibliography{reference}


\end{document}